\documentclass[a4paper,conference]{IEEEtran}
\IEEEoverridecommandlockouts

\usepackage{amsmath,graphicx,gensymb}
\usepackage{subcaption}
\usepackage{amssymb}

\newcommand{\etal}{\emph{et al.}}

\begin{document}

\title{Two-Level Attention-based Fusion Learning for RGB-D Face Recognition}


\author{\IEEEauthorblockN{Hardik Uppal, Alireza Sepas-Moghaddam,
Michael Greenspan, and Ali Etemad}
\IEEEauthorblockA{Department of Electrical and Computer Engineering \& Ingenuity Labs, Queen's University, Kingston, Ontario, Canada}
}

\maketitle

\begin{abstract}
With recent advances in RGB-D sensing technologies as well as improvements in machine learning and fusion techniques, RGB-D facial recognition has become an active area of research. A novel attention aware method is proposed to fuse two image modalities, RGB and depth, for enhanced RGB-D facial recognition. The proposed method first extracts features from both modalities using a convolutional feature extractor. These features are then fused using a two layer attention mechanism. The first layer focuses on the fused feature maps generated by the feature extractor, exploiting the relationship between feature maps using LSTM recurrent learning. The second layer focuses on the spatial features of those maps using convolution. The training database is preprocessed and augmented through a set of geometric transformations, and the learning process is further aided using transfer learning from a pure 2D RGB image training process. Comparative evaluations demonstrate that the proposed method outperforms other state-of-the-art approaches, including both traditional and deep neural network-based methods, on the challenging CurtinFaces and IIIT-D RGB-D benchmark databases, achieving classification accuracies over $98.2\%$ and $99.3\%$ respectively. The proposed attention mechanism is also compared with other attention mechanisms, demonstrating more accurate results. 
\end{abstract}

\begin{IEEEkeywords}
Facial Recognition, Attention, Modality Fusion, RGB-D images
\end{IEEEkeywords}

\section{Introduction}
\label{sec:intro}
Facial recognition (FR) technology has improved significantly over the past decade, both due to recent advancements in deep learning methods \cite{masi2018deep}, as well as its utility and wide-spread application in authentication processes in mobile phones and other electronic devices \cite{faceS}. Following the introduction of AlexNet in 2012~\cite{23}, most FR tasks such as face verification (one-to-one) and face identification (one-to-many), have employed a deep neural network approach with a CNN as the backbone \cite{masi2018deep}. Most of these advancements make use of 2D RGB (Red, Green and Blue channel) images, which are plentiful and readily available, thereby facilitating the training of even deeper and more effective neural networks. 

While CNNs have proven successful with RGB images \cite{masi2018deep}, recognition tasks using RGB-D images, which comprise co-registered color and range (depth) information, have been less thoroughly explored. The advent of inexpensive depth sensors such as the Microsoft Kinect and the Intel RealSense has reduced the cost of acquiring RGB-D images \cite{ zhang2012microsoft,Keselman_2017_CVPR_Workshops}. Depth (range) information can be instrumental for FR, as it provides geometric information about the face, in the form of dense 3D points that sample the surface of facial components~\cite{RGBDFR}. This additional information has shown to improve the effectiveness of FR algorithms, making them more accurate and robust to variations in pose and illumination \cite{PAMI,mian}. Traditional RGB-D approaches to FR used hand-crafted descriptors for the RGB and depth modalities to perform classification \cite{mian}, but these engineered features may not generalize well to all datasets. 

Generally, deep learning approaches for RGB-D facial recognition use different multimodal learning strategies such as feature-level or score-level fusion \cite{lee2016accurate}, following CNN feature extraction. Nonetheless, different parts of the embeddings and different input modalities may contain varying amounts of identity-related information, which most current fusion strategies fail to exploit. This is due to the fact that existing fusion strategies apply similar importance to different modalities and parts of the learned embedding. Attention mechanisms have recently achieved impressive results by focusing on certain parts of embeddings in various other complicated tasks such as image captioning and visual question answering. These mechanisms can be integrated into deep convolution networks to selectively learn the importance of feature embeddings, with respect to each other, in an end-to-end trainable fashion.

Accordingly, we propose here a novel method to effectively fuse the two RGB and depth modalities using an attention mechanism. The first level, called \textit{feature-map attention}, selectively focuses on the fused feature maps produced by the convolution layers using LSTM recurrent learning. The second level, called \textit{spatial attention}, selectively focuses on spatial convolutional information over the feature maps. The attention refined features are then further learned by fully connected layers for classification. The efficiency of our proposed attention-based fusion method has been tested on two RGB-D face recognition datasets, comprising CurtinFaces and IIIT-D RGB-D, using different and challenging test protocols. The experimental results showed the proposed method to have very promising performance, outperforming the baseline and state-of-the-art RGB-D face recognition methods. 

Our main contributions are as follows:
\begin{itemize}
\item We introduce a novel multimodal fusion mechanism for RGB-D FR using attention to selectively learn useful information from both RGB and depth modalities;
\item We perform ablation experiments on a number of variations of attention feature and spatial mechanisms, and demonstrate the performance improvement in attention-based fusion of the two modalities;
\item Our proposed method outperforms several other methods, setting new state-of-the-art values on two public RGB-D face datasets.
\end{itemize}

The rest of the paper is organized as follows: Section~\ref{sec:Related work} provides an overview on RGB-D face datasets, recognition methods, and attention mechanisms. Our proposed attention-based fusion solution is presented in Section~\ref{sec:Method}. Section~\ref{sec:Experiment} presents the experimental setup, implementation details, and performance comparison with other state-of-the-art RGB-D FR methods. Section~\ref{sec:ablation} presents the ablation experiments and analysis conducted on our network. Finally, Section~\ref{sec:Conclusion} concludes the paper with a discussion of future work.

\section{Related Work}
\label{sec:Related work}
\subsection{RGB-D Face Recognition Datasets and Their Benchmarks}
Given the need for specialized cameras for capturing RGB-D images, a few RGB-D datasets are available in the literature, which will be reviewed here. Hg \etal~\cite{4} collected the VAP dataset, containing 1149 images from 31 subjects, proposed a face detection algorithm based on curvature analysis. Min \etal~\cite{5} developed the EURECOM RGB-D face database and applied various algorithms to their baseline results on this dataset using a combination of Principal Component Analysis (PCA), Local Binary Patterns (LBP), and Scale Invariant Feature Transform (SIFT) features. Goswami \etal~\cite{1,2} collected the IIITD RGB-D dataset(4605 images), and then concatenated Histograms of Oriented Gradients (HOG) features extracted from saliency and entropy maps, and used a random forest as a classifier. Zhang \etal~\cite{7} developed the BUAA Lock3DFace RGB-D face database containing 5711 videos on 509 subjects and provided baseline results for the depth modality using the Iterative Closest Point (ICP) algorithm. CurtinFaces is another notable RGB-D dataset collected by Li \etal~\cite{mian}, which contains 5000 images for both modalities recorded using Kinect sensors. Sepas-Moghaddam \etal~\cite{LFFD} developed the IST-EURECOM light field face dataset containing RGB multi-view information, from which depth information was extracted. FaceWarehouse~\cite{FaceWarehouse} are specifically designed for facial emotion recognition, which prevents their usage in our experiments. Most of the above RGB-D datasets normally have thousands of training images as opposed to RGB datasets like VGGFace2, which has more than million images available.

\subsection{Traditional RGB-D Approaches}
Goswami \etal~\cite{1,2} used different types of features, such as Visual Saliency Map (VSM) from RGB data and entropy maps from depth data, which were then fused together with HOG features of image patches and fed to a classifier. In later work, they improved the feature set with RGB-D Image Saliency and Entropy maps (RISE) and Attributes from Depth Maps (ADM). Li \etal~\cite{mian} used various pre-processing methods to exploit face symmetry at the 3D point cloud level and obtained a frontal view, shape and texture irrespective of the pose. They used pose correctness with a Discriminant Color Space (DCS) transformation to improve the accuracy of their approach. Hayat \etal~\cite{29} used a co-variance matrix representation on the Riemannian manifold to represent the images and used a SVM classifier with a score level fusion method to fuse the depth and RGB scores to classify identities.

\subsection{Deep Learning Approaches}
Socher \etal~\cite{10} proposed a Convolutional Recursive Neural Network (CRNN) for RGB-D object recognition. In this network, two CNN networks were trained separately on the RGB and depth images, and the learned embeddings were fed into two RNN networks to obtain compositional features and part interactions. Borghi \etal~\cite{11} trained a Siamese CNN on RGB and depth images for a facial verification task~\cite{12}. Chowdhury \etal~\cite{3} built upon the work of Goswami \etal~\cite{1,2}, with an approach called learning based reconstruction. They used Autoencoders to get the mapping function from RGB and depth images, and used the reconstructed images from the mapping function for identification. 
Zhang \etal~\cite{13} addressed feature fusion using deep learning techniques, focusing on jointly learning the CNN embedding to fuse the common and complementary information offered by the two modalities together effectively.

\subsection{Attention in Images}
The attention mechanism has been shown to be effective in understanding images ~\cite{15,16,17}, and has been widely used in various tasks, including machine translation \cite{14}, visual question answering \cite{15,xu2016ask,nam2017dual}, object detection \cite{16}, semantic segmentation \cite{17} and person re-identification \cite{18}. Bottleneck Attention Module (BAM) was proposed by Park \etal~\cite{park2018bam} which introduced a light module that can be integrated with any feed-forward CNN. The module calculates an attention map along the channel and spatial dimensions. The module is specifically placed in the bottleneck of the network where down sampling of features is taking place, and is able to construct a hierarchical attention process.

Convolutional Block Attention Module (CBAM) is another example of using an attention mechanism for images. In this work Woo \etal~\cite{19} focus on building a lightweight attention module using both spatial attention and channel attention. This module is compatible with the convolution layers and is a very lightweight module which could be added to most CNNs with negligible overheads. CBAM has also been used for fusion for RGB-D face images for presentation attack detection task in~\cite{wang2019multi}. 
An LSTM network with convolutional features was used to exploit the spatial information in various image captioning tasks to generate attention weights for annotation vectors to caption the images~\cite{xu2015show}.
Sepas-Moghaddam \etal~\cite{Alireza2019attention}  used bidirectional LSTM (BLSTM) to produce attention weights for some specific spatio-angular features from multi-view cameras. 

\begin{figure}[h]
  \centering
  \includegraphics[width=.8\linewidth]{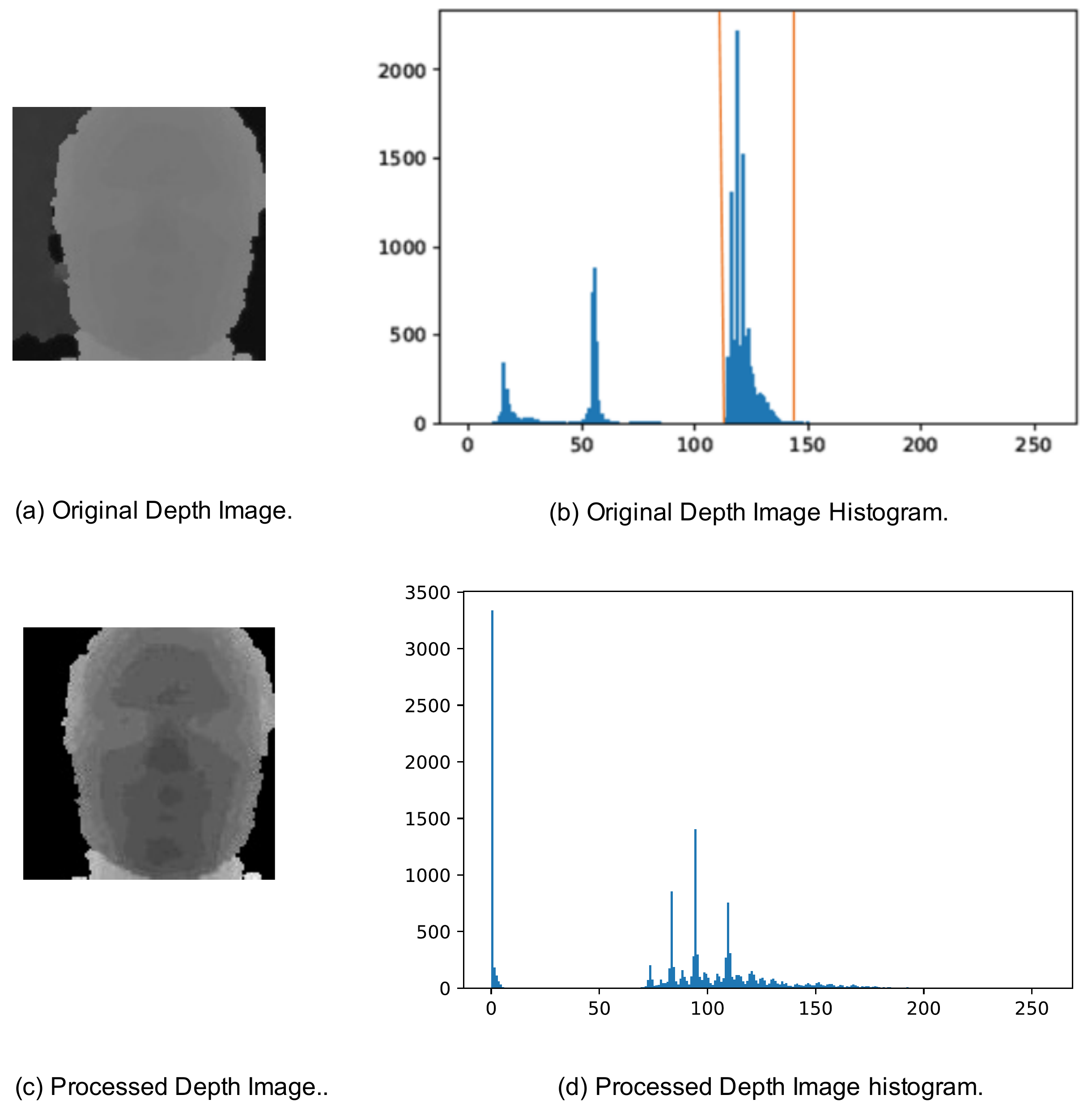}  
\caption{Depth image preprocessing. In Figure\ref{fig:fig_hist}(b) the orange lines represent the clipping planes, to clip out data points that are too close to camera or very far from it.}
\label{fig:fig_hist}
\end{figure}

\begin{figure*}[!t]
\centering
\includegraphics[width=1\linewidth]{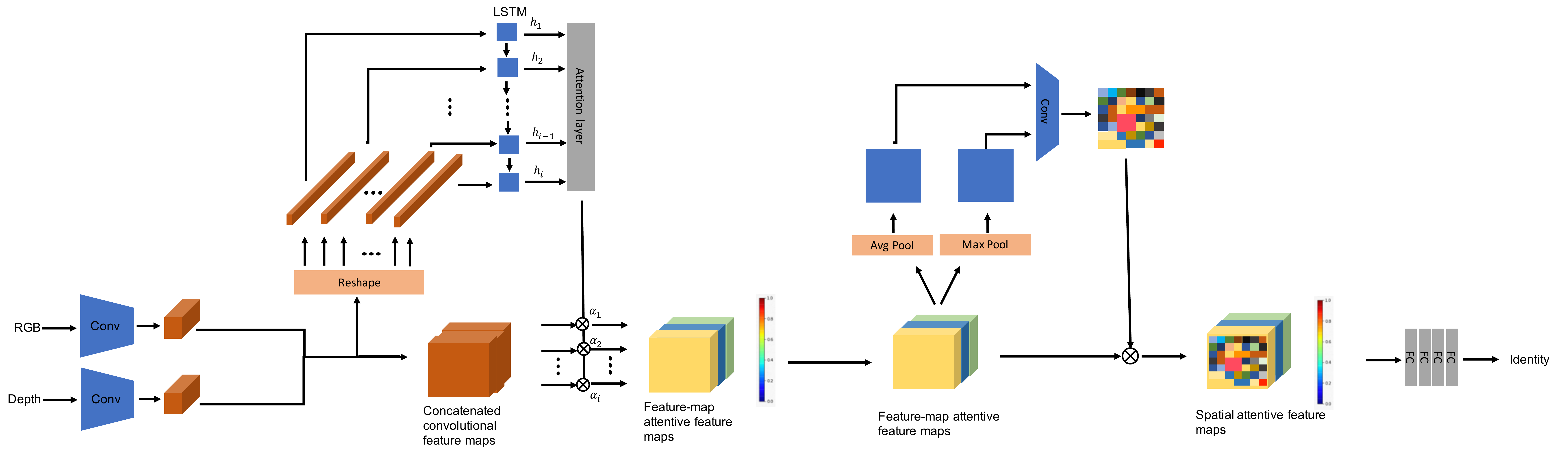}
\caption{Architecture of proposed two-level attentive network.}
\label{fig:architecture}
\end{figure*}

\section{Method}
\label{sec:Method}
We aim to develop an accurate FR method that is more robust to variations in environmental illumination and face pose. To this end, we present a multimodal recognition method using both RGB and depth modalities contained in Kinect images. We fuse the two modalities together using two attention mechanisms as depicted in Figure~\ref{fig:architecture}.

\subsection{Preprocessing and Image Augmentation}
\label{sec:augmentation}
The first preprocessing step is to determine two depth values
that respectively represent the near and far clipping planes (orange lines) of the scene as shown in Figure~\ref{fig:fig_hist}. The near and far clipping planes  in
Figure~\ref{fig:fig_hist}(b) are calculated by only keeping the depth values between the \(90^{th}\) and the  \(25^{th}\) percentile value, which were determined empirically. These clipping planes are used to filter out content that are too near or too far from the camera, passing through the face information. After this process we normalize the remaining content to fall within the 0 to 255 values, to make full use of the dynamic range.

Following this preprocessing, both the RGB and the (preprocessed) depth images are passed through the dlib CNN face extractor network~\cite{20}  
which returns cropped images containing only face regions. To combat the small size of RGB-D databases and to make the model  more robust and less sensitive to small changes in the images, we apply image augmentation to the dataset during training. For augmentation, we use the geometric transformations listed in Table~\ref{tab:augmentation}, which include image rotation, mirror reflection (flip about the vertical axis), as well as affine (sheer) and perspective (warp) transformations. This increases our dataset size to 4 times its original size as the augmentation is applied to each image.

\begin{table}[h]
\caption{Image Augmentation}
\small
    \label{tab:augmentation}
    \centering
    \begin{tabular}{l l c}
    \hline
     \textbf{Attribute} & \textbf{Parameter}  & \textbf{Value} \\
     \hline
     Rotation &  Angle & -30\degree\  to 30\degree\\
     Sheer & Angle & -16\degree\ to 16\degree\\
     Flip & Axis & Vertical\\
     Perspective & Scale & 50\% to 150\%\\
    \hline
    \end{tabular}
\end{table}

\subsection{Network Architecture}
\label{sec:architecture}
Our network consists of a Siamese convolution network unit for RGB and depth modalities following the architecture of the VGG network~\cite{21}. We utilize the convolutional feature extraction part of the VGG network which has already been trained on over 3.3+ million images from the VGGFace2 dataset \cite{28}. This helps to speed up the training process, by transferring the network's already learnt ability to identify facial features, and also to further compensate for the relatively small size of the RGB-D datasets. Both depth and RGB images are passed through the convolutional part of the network to obtain feature embeddings, in the form of convolutional tensors, for both modalities from the 13\textsuperscript{th} layer of the VGG-16 network. 

Following convolutional feature extraction, the next part of the network is the attention mechanism, which focuses the network on the salient parts of the feature embeddings. The attention mechanism is partitioned into two different units, named \emph{feature-map attention} and \emph{spatial attention}.

\subsubsection{Feature-Map Attention}
The aim of this mechanism is to help train the network to focus on those feature maps generated in the embedding that have a greater contribution~\cite{chen2017sca,zhang2019classification,zhang2019capsule, Alireza2019attention} towards our classification task. 
Here, an LSTM layer first acts as the conditional encoder for each feature map to then calculate the attention weights using the subsequent dense layer.
In this context, we concatenate the features extracted by the CNN networks from both RGB and depth modalities as shown in Figure~\ref{fig:architecture}. We divide the concatenated convolutional tensor into feature maps to feed the LSTM layer whose number of inputs equals the number of convolutional feature maps as listed in Table~\ref{tab:parameter}. The LSTM layer encodes the captured information for each map and allows the network to keep context of all the maps whilst calculating attention. The output of the LSTM layer feeds a dense attention layer with sigmoid activation, thus computing attention weights for map in a range between 0 and 1. This assigns higher attention weights to the more informative maps. We attempted to tackle this problem by using different attention mechanisms also, and that comparison has been described in Section~\ref{sec:ablation_candidates}.

Let \textbf{\(F_{RGB}\)} and \textbf{\(F_{Depth}\)} be the feature embeddings extracted from the CNN network, where \(F_{RGB}\) and  \(F_{Depth} \in \mathbb{R}^{M \times M \times K}\) and $K$ is the number of channels of convolutional feature maps. \(F_{concat}\) is then given by:
\begin{equation}
 F_{concat} = F_{RGB}\oplus F_{Depth}
 \label{eq:fconcat}
\end{equation}
\noindent where \(\oplus\) refers to the concatenation operation and
\(F_{concat} \in \mathbb{R}^{M \times M \times C}\) and $C=K+K$, where $C$ is number of fused convolutional feature maps and $M$ is spatial size of the convolutional feature maps. These concatenated features are reshaped to $C$ partial vectors, $F_{pv}$, to feed to the LSTM layer, where \(F_{pv} \in \mathbb{R}^{M^{2} \times 1}\). Each LSTM output $h_{i}$, where $i$ is the $i^{th}$ $F_{pv}$ vector,
is fed to the attention layer with trainable attention weights $W_{fm}$ for each feature map of $F_{concat}$ as:
\begin{equation}
\begin{aligned}
\theta_{fm} = (W_{0}h_{i} + b_{0})
\end{aligned}
\label{eq:Tatt1}
\end{equation}
\begin{equation}
\begin{aligned}
W_{fm} = \frac{\mathrm{1} }{\mathrm{1} + e^{- \theta_{fm}^T}}
\end{aligned}
\label{eq:Watt1}
\end{equation}
\noindent where \(\theta_{fm} \) is the output of the multi-level perceptron; \(W_{0}\) are the trainable weights learnt in the attention layer, and; \(b_{0}\) is the respective biases. This output is normalized to $[0,1]$ with the sigmoid function in Eq.~\ref{eq:Watt1}. 
The feature-map attention refined features \(F_{fm}\) are then calculated as:
\begin{equation}
F_{fm} = W_{fm} \times F_{concat}
\end{equation}

\subsubsection{Spatial Attention}
After refining the features through feature-map attention, the network next focuses attention on the spatial axis of the embedding. This module helps the network to identify the most salient features in the embedding and to focus its attention on those features \cite{19}. To take the most salient information from the feature embedding, we use average and max pooling along the feature map axis. To obtain the attention weight we pass these average \(F_{avg1}\) and max pooled \(F_{max1}\) features to a convolution layer with kernel size of $1\times1$ and $1$ feature map to calculate the weights to a single weight layer, followed by sigmoid activation as shown in Figure~\ref{fig:architecture}. Performing average pooling and max pooling on the \(F_{fm}\) feature embedding results in \(F_{avg1}\) and \(F_{max1}\). The spatial attention weights \(W_{spatial}\) are then calculated as:
\begin{equation}
\begin{aligned}
\theta_{spatial} = Conv(F_{avg1}\oplus F_{max1})
\end{aligned}
\end{equation}
\begin{equation}
\begin{aligned}
W_{spatial} = \frac{\mathrm{1} }{\mathrm{1} + e^{- \theta_{spatial}^T }}
\end{aligned}
\end{equation}

The final refined features following the attention module are then given by:
\begin{equation}
F_{attention} = W_{spatial} \times F_{fm}
\end{equation}

After fusing these features together, a classifier is applied to segregate the features obtained from the attention layer. We use 4 fully connected layers to serve as a classifier, with 3 layers followed by batch-norm and dropout to regularize the output. The final layer is the dense layer with the number of hidden units equal to the number of classes.  

\begin{table}[h!]
\centering
\caption{The optimal values obtained for the proposed attention-based fusion.}
\setlength
\tabcolsep{4pt}
\begin{tabular}{ l| l| l}
\hline
\textbf{Module} & \textbf{Parameter} & \textbf{Setting} \\
\hline
\hline
    Convolutional feature  & Architecture   &  VGG-16  \\
    extractor  & Pre-trained weights & VGG-Face2 \\
      & Convolution feature size($F_{RGB}$) & $7\times7\times512$ \\
      & Convolution feature size($F_{D}$) & $7\times7\times512$ \\

    \hline

    Feature-map attention &  Input image feature size &  $7\times7\times1024$  \\
      & Output attention map  & $1\times1\times1024$ \\    
      & Reshape & $49\times1024$ \\
      & LSTM layer size & $1024$ \\
      & Attention layer size & 1024 \\
      & Attention layer activation & softmax \\
      & Input image feature size & $7\times7\times1024$ \\
    \hline

    Spatial attention &  Input feature size  & $7\times7\times1024$  \\
      & Output attention map  & $7\times7\times1$ \\
      & Convolution kernel size & 1 \\
      & Convolution feature maps & 1 \\
      & Convolution activation & softmax \\
    \hline
    
    Classifier &  Classifier layers  &  4 FC  \\
      & Number of layer 1 nodes & 2048 \\
      & Number of layer 2 nodes & 1024 \\
      & Number of layer 3 nodes & 512 \\
      & Dropout ratio & 0.5 \\
      & Number of layer 4 nodes & No. of classes \\
      & Layer 4 activation & softmax \\
    \hline

     Full network  & Batch size & 20 \\
      & Loss function & Cross entropy \\
      & Optimizer & Adam \\
      & Learning rate & 0.00001 \\
      & Learning rate decay & 0.9 \\
      & Metric & Accuracy \\
    \hline
\end{tabular}
\label{tab:parameter}
\end{table}

\begin{table*}[t]
\setlength{\tabcolsep}{6pt} 
\renewcommand{\arraystretch}{1} 
\centering
\caption{Performance comparison on the IIIT-D dataset.}
\vspace{2mm}
\begin{tabular}{ l l l l l l c }
  \hline
  \textbf{Ref.} & \textbf{Year} & \textbf{Authors} & \textbf{Feat. Extractor} & \textbf{Classifier}& \textbf{Input} & \textbf{Accuracy} \\
  \hline
    \cite{25}& 2016 & He \etal\ &ResNet-50&FC/Softmax& RGB & 95.8\%
  \\
\cite{hu2018squeeze}& 2017 & Hu \etal\ &SE-ResNet-50&FC/Softmax& RGB & 96.4\% 
  \\  
  \cite{1}& 2013 & Goswami \etal\ & RISE & Random Forest& RGB + Depth & 91.6\%  \\
  \cite{2}& 2014 & Goswami \etal\ & RISE & Random Forest& RGB + Depth & 95.3\%  \\
  \cite{13}& 2018 & Zhang \etal\ &9 Layers CNN + Inception &FC/Softmax& RGB + Depth  & 98.6\%  \\
  \cite{3}& 2016 & Chowdhury \etal\ & Autoencoder & FC/Softmax & RGB + Depth & 98.7\% \\
  & 2020 & \textbf{Proposed} &VGG + Two-level Attention & FC/Softmax& RGB + Depth  & 99.4\% \\
  \hline
\end{tabular}
\label{tab:IIIT-D}
\end{table*}

\begin{table*}[t]
\setlength{\tabcolsep}{6pt} 
\renewcommand{\arraystretch}{1} 
\centering
\caption{Performance comparison on the CurtinFaces dataset.}
\vspace{2mm}
\begin{tabular}{ l l l l l l c c  c}
  \hline

\multicolumn{6}{l}{\textbf{}} & \multicolumn{3}{c}{\textbf{Accuracy}}  \\
 \cline{7-9}
  \textbf{Ref.} & \textbf{Year} & \textbf{Authors} & \textbf{Feat. Extractor} & \textbf{Classifier}& \textbf{Input}  & \textbf{Pose} & \textbf{Illumination} & \textbf{Average} \\

  \hline
    \cite{25}& 2016 & He \etal\ &ResNet-50&FC/Softmax& RGB & 94.4\% & 96.0\%& 95.7\% 
  \\
\cite{hu2018squeeze}& 2017 & Hu \etal\ &SE-ResNet-50&FC/Softmax&  RGB & 97.4\% & 98.2\%& 97.8\% 
  \\
  \cite{mian}& 2013 &  Li \etal\ & Discriminat Color Space Trans. & SRC & RGB + Depth  & 96.4\% & 98.2\%& 97.3\% 
  \\
  \cite{li2016face}& 2016 & Li \etal\ & LBP + Haar + Gabor& SRC &  RGB + Depth & -- & --  & 91.3\% 
  \\
  \cite{29}& 2016 & Hayat \etal\ & Covariance Matrix Rep. & SVM & RGB + Depth  & -- & --& 96.4\%  
  \\
  & 2020 & \textbf{Proposed} &VGG + Two-level Attention &FC/Softmax& RGB + Depth  & 97.5\% & 98.9\%& 98.2\% \\
  \hline
\end{tabular}
\label{tab:CurtinFaces}
\end{table*}
\section{Experiments}
\label{sec:Experiment}
\subsection{Datasets}
\textbf{IIIT-D RGB-D }\cite{1}:
This dataset 
contains images from 106 subjects captured using a Microsoft Kinect. Each subject has multiple images ranging from 11 to 254, with $640\!\!\times\!\!480$ resolution. The dataset already has a pre-defined protocol with a five-fold cross-validation strategy, to which we strictly adhered in our experiments. Each of the five folds has 4 images per subject for training and 17 images for testing.

\noindent\textbf{CurtinFaces  RGB-D} \cite{mian}: The CurtinFaces dataset 
contains over 5000 images of 52 subjects with both RGB and depth modalities, captured with a Microsoft Kinect. For each subject, the first 3 images are the frontal, right and left poses. The remaining 49 images comprise 7 different poses recorded with 7 different expressions, and 35 images in which 5 different illumination variations are acquired with 7 different expressions. It also contains images with sunglasses and hand occlusions. We follow the test protocol described in ~\cite{mian}.
There are a total of 87 images per subject, from which 18 are used for training, 30 make up test set 1 which contains pose and expression variation, and 39 make up test set 2 which contains expression and illumination variation.

\begin{figure}[h!]
\centering
\includegraphics[width=1\linewidth]{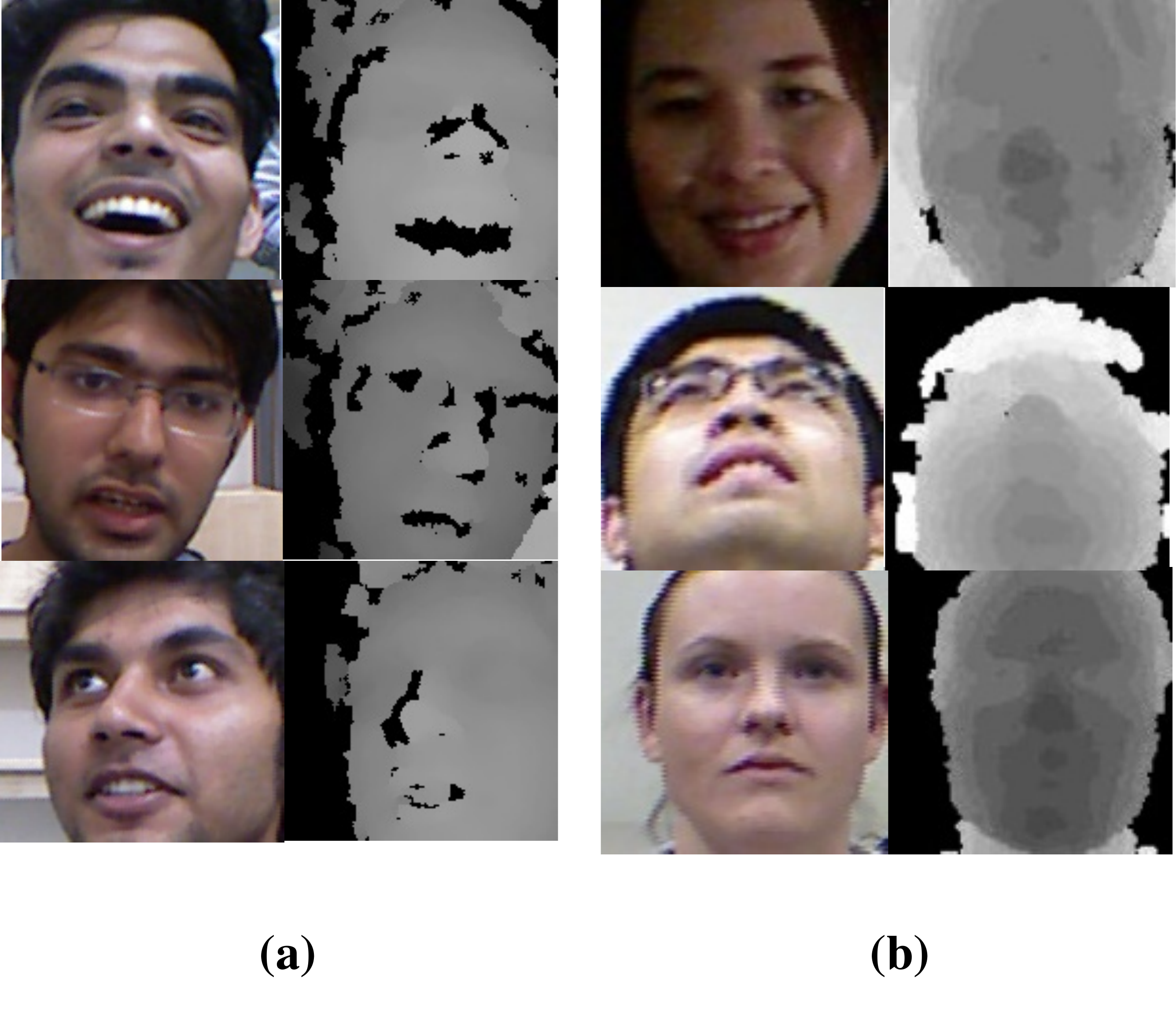}
\caption{Sample images from datasets; (a) IIIT-D RGB-D dataset and (b) CurtinFaces dataset.}
\label{fig:db_icpr}
\end{figure}




\subsection{Implementation} 
The convolution layers were initialized with weights from the VGGFace2 dataset \cite{28}. We used the Adam optimizer with a learning rate of 0.00001 and a decay rate of 0.9. The dropout rate of 0.5, size of three classifier nodes as 2048, 1024, and 512 respectively, and batch size of 20 were determined by grid search. The input to the network was the synchronized RGB and depth images, after image augmentation as described in Section \ref{sec:augmentation}. The two attention modules described in Section \ref{sec:architecture} were applied to the concatenated features from the convolution layers, and the attention refined features were fed to the classifier network containing 3 fully connected layers with 2048, 1024 and 512 nodes respectively. The fourth and final fully connected layer comprised the  dataset classes. The implementation used Python3.6 and Keras \cite{keras} with a TensorFlow \cite{tensorflow} back end to develop our model, and used data generators from Keras to setup the flow of the data pipelines to the model. The proposed deep network was trained using a Nvidia GTX 1070 GPU. All the optimum parameters are mentioned in Table~\ref{tab:parameter} for reproducibility of results.

\begin{table*}[!h]
\renewcommand{\arraystretch}{1.2} 
\centering
\caption{Ablation study on the two datasets.}
\vspace{2mm}
\begin{tabular}{ l | c |c c c}
 \hline
\multicolumn{1}{l}{\textbf{}} &  
\multicolumn{1}{|c}{\textbf{IIIT-D RGB-D}} &
\multicolumn{3}{|c}{\textbf{CurtinFaces}}
\\
 \cline{2-5}
  \textbf{Model} & \textbf{Average}& \textbf{Pose} & \textbf{Illumination} & \textbf{Average}\\
  \hline
  VGG-Face (RGB) & 94.1\%& 92.5\% & 93.2\% & 92.8\%\\
  VGG-Face (Depth)  & 68.5\% & 60.2\% & 63.2\%& 61.7\% \\
  VGG-Face (RGB $\oplus$ Depth) & 95.4\% & 92.6\% & 94.2\% & 93.4\% \\
  With Feature-Map Attention & 96.4\ & 97.5\% & 98.3\% & 97.8\% \\
  With Spatial Attention &  96.2\% &  96.7\% & 97.3\% & 97.0\% \\
  \textbf{Two-level Attention} & \textbf{99.4}\% & \textbf{98.1}\% & \textbf{99.1}\% &\textbf{98.6}\% \\
  \hline
\end{tabular}
\label{tab:all_ablation}
\end{table*}

\begin{table*}[t]
\renewcommand{\arraystretch}{1.2} 
\centering
\caption{Comparison of accuracy with different attention mechanisms on both datasets.}
\vspace{2mm}
\begin{tabular}{ l l | c |c c c }
\hline
\multicolumn{1}{l}{\textbf{}} & 
\multicolumn{1}{l}{\textbf{}} & 
\multicolumn{1}{|c}{\textbf{IIIT-D RGB-D}} &
\multicolumn{3}{|c}{\textbf{CurtinFaces}}
\\
 \cline{3-6}
  \textbf{Candidate}&\textbf{Attention Type} & \textbf{Average} &\textbf{Pose}& \textbf{Illumination}& \textbf{Average}\\
  \hline
   Dense Layer & Feature-Map Attention &96.9\%&96.7\%& 98.9\%& 97.8\% \\
   LSTM + Dense Layer & Feature-Map Attention &98.5\%& 97.8\% & 99.1\%& 98.4\%\\
   Dense Layer & Spatial Attention &96.4\% &96.2\%& 96.9\%& 96.5\% \\
   Convolution Layer & Spatial Attention &97.8\%&96.7\%& 97.3\%& 97.0\% \\
   \textbf{LSTM + Dense and Convolution}& \textbf{Two-level Attention} &\textbf{99.4}\%& \textbf{98.1}\% & \textbf{99.1}\%& \textbf{98.6}\% \\
  \hline
\end{tabular}
\label{tab:comp_att_both}
\end{table*}

\subsection{Performance and Comparison}
To validate our results, we compared the performance of our proposed method with benchmark results from the CurtinFaces and IIIT-D RGB-D datasets. The results for the IIIT-D RGB-D dataset are shown in Table~\ref{tab:IIIT-D}.
Our proposed method increases the identification rank-1 accuracy to 99.38\%. The proposed method outperformed the state-of-the-art results by Chowdhury \etal~\cite{3} which uses depth rich features acquired from an autoencoder, achieving a classification accuracy of 98.7\%, and Zhang \etal~\cite{7}, which uses complimentary feature learning to achieve 98.6\% accuracy.

To further verify our results, we tested our multimodal attention network on the  CurtinFaces RGB-D dataset, the results of which are tabulated in Table~\ref{tab:CurtinFaces}. Our proposed model outperforms the state-of-the-art results in both the test sets, with 97.53\% accuracy in the pose-expression variation test set, and with 98.88\% accuracy in the illumination-expression variation test set.

\subsection{Ablation Experiments}
\label{sec:ablation}

\subsubsection{Ablation with the Architecture modules}
We employ our two attention modules on top of the convolution layers for effective fusion of the two modalities. To demonstrate the effectiveness of the two modules we conducted the  experiments described below, with results compiled in Table~\ref{tab:all_ablation} , respectively for IIIT-D and CurtainFaces datasets. It is evident from the results that the attention module aided with the effective fusion of the two modalities and improved overall performance. We also employed the attention modules separately to observe the improvement over the unaltered VGG network. 
It can be seen that employing feature-map attention alone improves accuracy by $1\%$, whereas when employing spatial attention alone, the accuracy is improved by $1.3\%$. The best performance was achieved by employing both attention mechanisms, yielding a considerable performance improvement for both datasets.

\subsubsection{Ablation within Attention modules}
\label{sec:ablation_candidates}
We further investigated the various attention mechanisms for feature-map and spatial attention modules listed below, and compare their performance with our attention modules in Table\ref{tab:comp_att_both} for IIIT-D RGB-D dataset and for CurtinFaces dataset:
\begin{itemize}
    \item Feature-Map Attention with Dense Layer: This module is inspired by channel attention used in \cite{19}. It uses a Dense layer with the number of neurons equal to the number of feature-maps present after concatenating the feature-maps extracted from the convolutional feature extractor from both modalities. Each neuron is tasked with calculating the attention weights for each map.
    \item Feature-Map Attention with LSTM + Dense Attention Layer: This module is built on top of the previous candidate. We add an LSTM layer to recurrently learn the feature maps in encoded form for which attention weights are calculated using Dense layer. We further performed experiments with different numbers of LSTM layers as well as a BLSTM  network architecture for feature-map attention as shown in Table~\ref{tab:Comp_lstm}. The results show that one and two LSTM layers deliver almost the same average performance, whereas 3 LSTM layers and a BLSTM network architecture slightly decrease the performance. The lower computational complexity justifies the selection of one LSTM layer in our final solution.    
    \item Spatial Attention with Dense Layer: We use a Dense layer with the same size as the spatial dimension of the feature embedding. Every neuron attempts to calculate the attention weight for spatial elements of the feature representation.
    \item Spatial Attention with Convolution Layer: We use convolutional layer with kernel size of 1 and 1 filter map to get the attention weights for each spatial element of the feature representation.
\end{itemize}

\begin{table}[h]
\setlength{\tabcolsep}{4pt} 
\renewcommand{\arraystretch}{1.2} 
\centering
\caption{Accuracy with layers of LSTM for feature-map attention.}
\vspace{2mm}
\begin{tabular}{ l | c |c c c }
\cline{1-5}
\multicolumn{1}{l}{\textbf{}} & 
\multicolumn{1}{|c}{\textbf{IIIT-D RGB-D}} &
\multicolumn{3}{|c}{\textbf{CurtinFaces}}
\\
 \cline{2-5}
  \textbf{Layers}  & \textbf{Average} & \textbf{Pose} & \textbf{Illum.} & \textbf{Average} 
  \\
  \hline
      1 LSTM layer & 98.5\% & 97.8\%  & 99.1\% & 98.4\%
  \\
    2 LSTM layers & 98.8\% &97.3\%  & 99.3\% & 98.3\% 
  \\
    3 LSTM layers & 98.1\% & 96.6\%  & 99.5\% & 98.0\% 
  \\
    1 BLSTM layer & 97.9\% & 96.4\%  & 99.0\% & 97.7\% 
  \\
  \hline
\end{tabular}
\label{tab:Comp_lstm}
\end{table}

\begin{figure*}[t]
\centering
\includegraphics[width=0.95\linewidth]{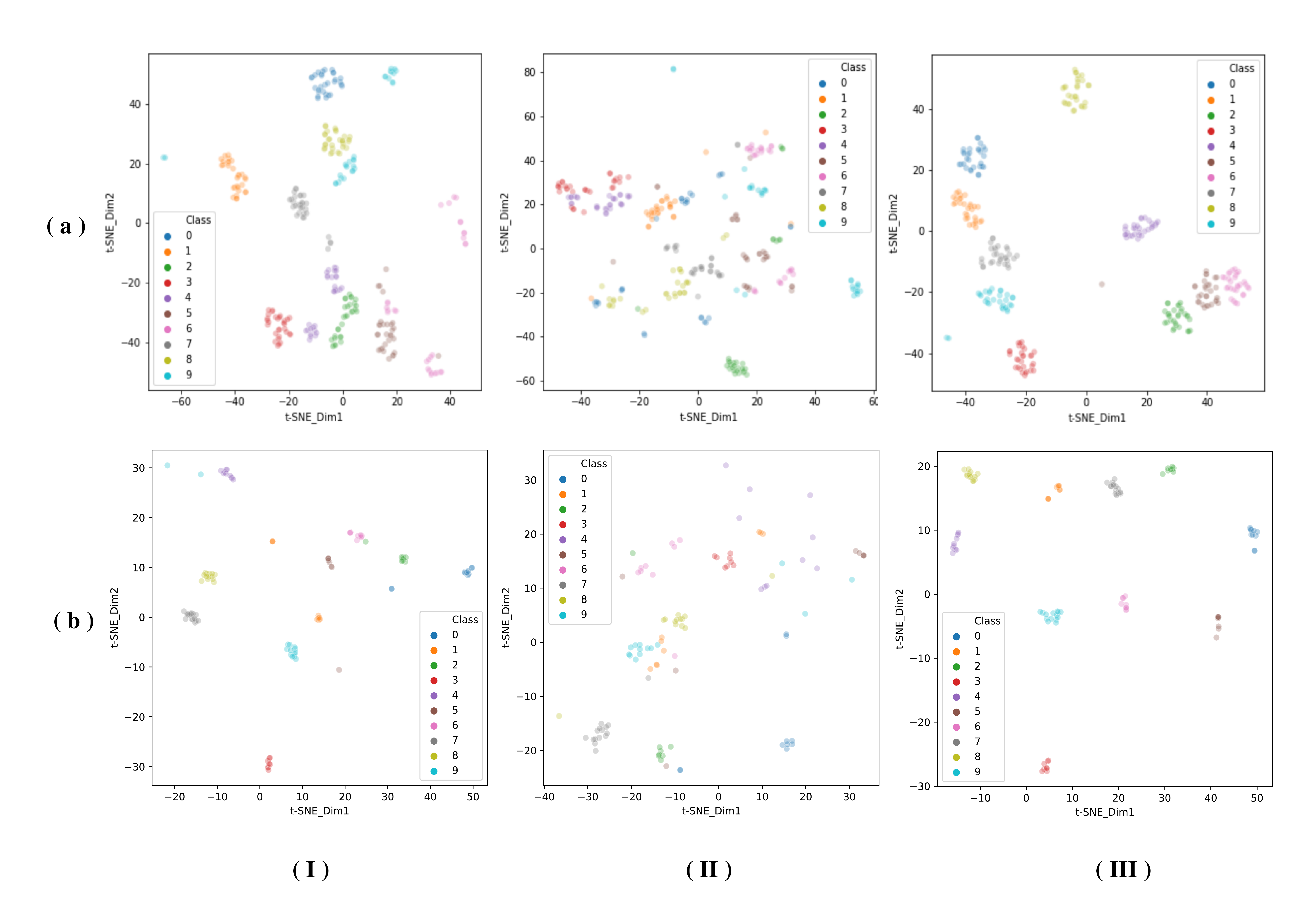}  
\caption{t-SNE visualization of our proposed method and other solutions. Every row in the figure represents a dataset, i.e. a) CurtinFaces and b) IIIT-D RGB-D.  Every column corresponds to a network and data modality, i.e.: (I) VGG with only RGB input, (II) VGG with only Depth input and (III)  attention-based fusion with RGB-D.}
\label{fig:tSne}
\end{figure*}

The results show that the feature map attention variation of LSTM + Dense layer performs significantly better than the Dense layer alone. For the spatial attention, the results show the superiority of the convolutional attention over the Dense layer for both datasets. Naturally, for our proposed method, we use the the best performing variations, i.e. LSTM + Dense attention layer for feature-map attention, and Convolution layer for spatial attention, which provide the best results in both of the RGB-D datasets.

\subsection{Analysis of Embedding with Attention}
The feature spaces that the embedding produced before and after the attention mechanism are illustrated in Figure~\ref{fig:tSne}. In particular, Figure~\ref{fig:tSne}~(III) shows the multimodal embedding produced after the attention mechanism, whereas Figure~\ref{fig:tSne}~(I) and Figure~\ref{fig:tSne}~(II) show the respective RGB and Depth embeddings individually. To aid this visualization, we use only 10 classes from the CurtinFaces dataset and 10 classes from IIIT-D RGB-D dataset. It can be observed that the multimodal embedding can produce more effective clustering that facilitates distinguishing between the classes. The RGB embedding alone is better than the Depth embedding alone, but both are not able to distinguish between a few classes. For instance, classes 2 and 4 have mixed clusters, whereas classes 5 and 6 have overlapping clusters. The best separation occurs with the attention-based fusion of the multimodal data, which produces the cleanest separation of all classes.

\section{Conclusion and Future Work}
\label{sec:Conclusion}
We presented an attention-based network to effectively fuse the RGB and depth modalities of RGB-D images for face recognition. Through our evaluations we validate
that our attention aware fusion offers more accurate rank-1 recognition results than the state-of-the-art methods on the IIIT-D RGB-D dataset at 99.4\% and on the CurtinFaces dataset at 98.6\%. We further investigate the performance of our attention mechanism in comparison with other attention mechanisms, where the results reveal the superiority of our method. We conduct extensive ablation trials, revealing the relative contributions of the individual components of our method towards the final performance.

In future work, we will explore the performance of our proposed solution using different sets of modalities such as thermal, speech, bio-signals. Moreover, we will explore Depth image generation from RGB images to aid the face recognition mechanism, as depth has been shown here to significantly affect the recognition accuracy.

\section{Acknowledgment}
The authors would like to thank Irdeto Canada Corporation and the Natural Sciences and Engineering Research Council of Canada (NSERC), for supporting this research.

\label{sec:ref}

\small
\bibliographystyle{IEEEbib}
\bibliography{bibfile}


\end{document}